\documentclass[10pt,twocolumn,letterpaper]{article}

\usepackage{cvpr}
\usepackage{times}
\usepackage{epsfig}
\usepackage{graphicx}
\usepackage{amsmath}
\usepackage{amssymb}

\usepackage{pifont}
\usepackage{bbding}
\usepackage{tabularray}
\usepackage{multirow}

\usepackage{booktabs}
\usepackage{xstring}
\usepackage{algorithm}
\usepackage{algpseudocode}

\newcommand{\formatword}[1]{\textbf{\textit{\MakeUppercase{#1}}}}
\usepackage[capitalize]{cleveref}
\Crefname{section}{Section}{Sections}
\Crefname{table}{Table}{Tables}
\crefname{figure}{Figure}{Figures}
\crefname{equation}{Equation}{Equations}

\begin{document}

\title{Adversarial Robustness in RGB-Skeleton Action Recognition: Leveraging Attention Modality Reweighter}

\author{Chao Liu$^{1}$ \quad\quad  Xin Liu$^2$\thanks{Xin Liu and Zitong Yu are corresponding authors} \quad\quad    Zitong Yu$^3$* \quad\quad    Yonghong Hou$^{1}$ \quad  \quad
Huanjing Yue$^{1}$ \quad  \quad Jingyu Yang$^{1}$ \\
$^1$Tianjin University \quad\quad
$^2$Lappeenranta-Lahti University of Technology \quad\quad
$^3$ Great Bay University  \\
{\tt\small xin.liu@lut.fi, yuzitong@gbu.edu.cn}
}

\maketitle
\thispagestyle{empty}

\begin{abstract}
   Deep neural networks (DNNs) have been applied in many computer vision tasks and achieved state-of-the-art (SOTA) performance. However, misclassification will occur when DNNs predict adversarial examples which are created by adding human-imperceptible adversarial noise to natural examples. This limits the application of DNN in security-critical fields. In order to enhance the robustness of models, previous research has primarily focused on the unimodal domain, such as image recognition and video understanding. Although multi-modal learning has achieved advanced performance in various tasks, such as action recognition, research on the robustness of RGB-skeleton action recognition models is scarce. In this paper, we systematically investigate how to improve the robustness of RGB-skeleton action recognition models. We initially conducted empirical analysis on the robustness of different modalities and observed that the skeleton modality is more robust than the RGB modality. Motivated by this observation, we propose the \formatword{A}ttention-based \formatword{M}odality \formatword{R}eweighter (\formatword{AMR}), which utilizes an attention layer to re-weight the two modalities, enabling the model to learn more robust features. Our AMR is plug-and-play, allowing easy integration with multimodal models. To demonstrate the effectiveness of AMR, we conducted extensive experiments on various datasets. For example, compared to the SOTA methods, AMR exhibits a 43.77\% improvement against PGD20 attacks on the NTU-RGB+D 60 dataset. Furthermore, it effectively balances the differences in robustness between different modalities. 
\end{abstract}

\section{Introduction}
Deep Neural Networks (DNNs) have achieved SOTA performance in various vision tasks, like segmentation \cite{1993A}, detection \cite{2016You}, and super-resolution \cite{2009Super}. However, DNNs are vulnerable to imperceptible adversarial perturbations\cite{Goodfellow2014ExplainingAH}. A finely crafted adversarial perturbation can easily fool the neural network. Adversarial attacks covered many deep learning tasks including video action recognition \cite{2018Sparse}, person re-identification \cite{2020Transferable}, and natural language processing \cite{2020BERT}. Given the DNN's widespread applications, especially in safety-critical domains like autonomous driving \cite{Chen2019ModelfreeDR,Eykholt2018RobustPA,lu2024gpt,zhang2024advancing} and medical diagnostics \cite{Buch2018ArtificialII,Kong2017CancerMD,Ma2019UnderstandingAA,liu2023multi,liu2024rppg}, enhancing neural network robustness is crucial.

\begin{figure}
    \centering
    \includegraphics[width=1\linewidth]{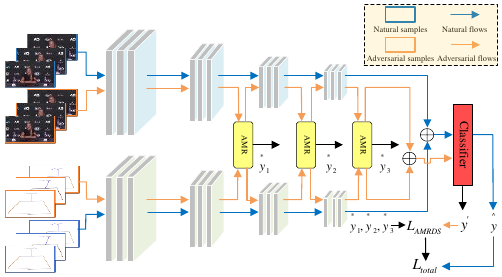}
    \caption{The framework of the proposed method AMR. During the training, we feed-forward both natural and adversarial samples from two modalities into the multimodal network in parallel. Features extracted from adversarial samples are input into AMR, and their weighted counterparts enter respective subsequent layers of the network. These features are simultaneously used for classification to generate auxiliary predictions denoted as $\overset{*}{\mathop{y}}\ $, contributing to the loss function. The natural samples go through regular forward propagation to yield prediction results.}
    \label{fig:framework}
\end{figure}

To address DNN vulnerability, efforts are categorized into adversarial attacks and defenses. Various methods are proposed to generate diverse adversarial samples \cite{2013Intriguing,2016Towards,2017Towards,2015DeepFool,Su2017OnePA}. Many works aim to defend against these attacks, with adversarial training showing the most reliable robustness. Adversarial training inputs adversarial examples into each training step. Natural examples are augmented with worst-case perturbations within a small $l_{p}$-norm ball, smoothing the loss landscape and improving classification performance in boundary regions.

As DNNs advances, multimodal technology has become a widely studied field with successful applications in vision-language navigation \cite{Wang2023ADS}, image-text matching\cite{Fu2023LearningSR}, and medical image diagnosis \cite{Khader2022MedicalDW,Moon2021MultiModalUA}. Multimodal technology offers advantages over unimodal technology. Previous research on model robustness has focused mainly on unimodal domains, like video comprehension \cite{Kinfu2022AnalysisAE}, with limited attention to multimodal models. Thus, research on enhancing the robustness of multimodal models is urgently needed. This paper systematically analyzes the robustness of action recognition models based on RGB-skeleton data and proposes methods to enhance their robustness.

\begin{figure}
    \centering
    \includegraphics[width=1\linewidth]{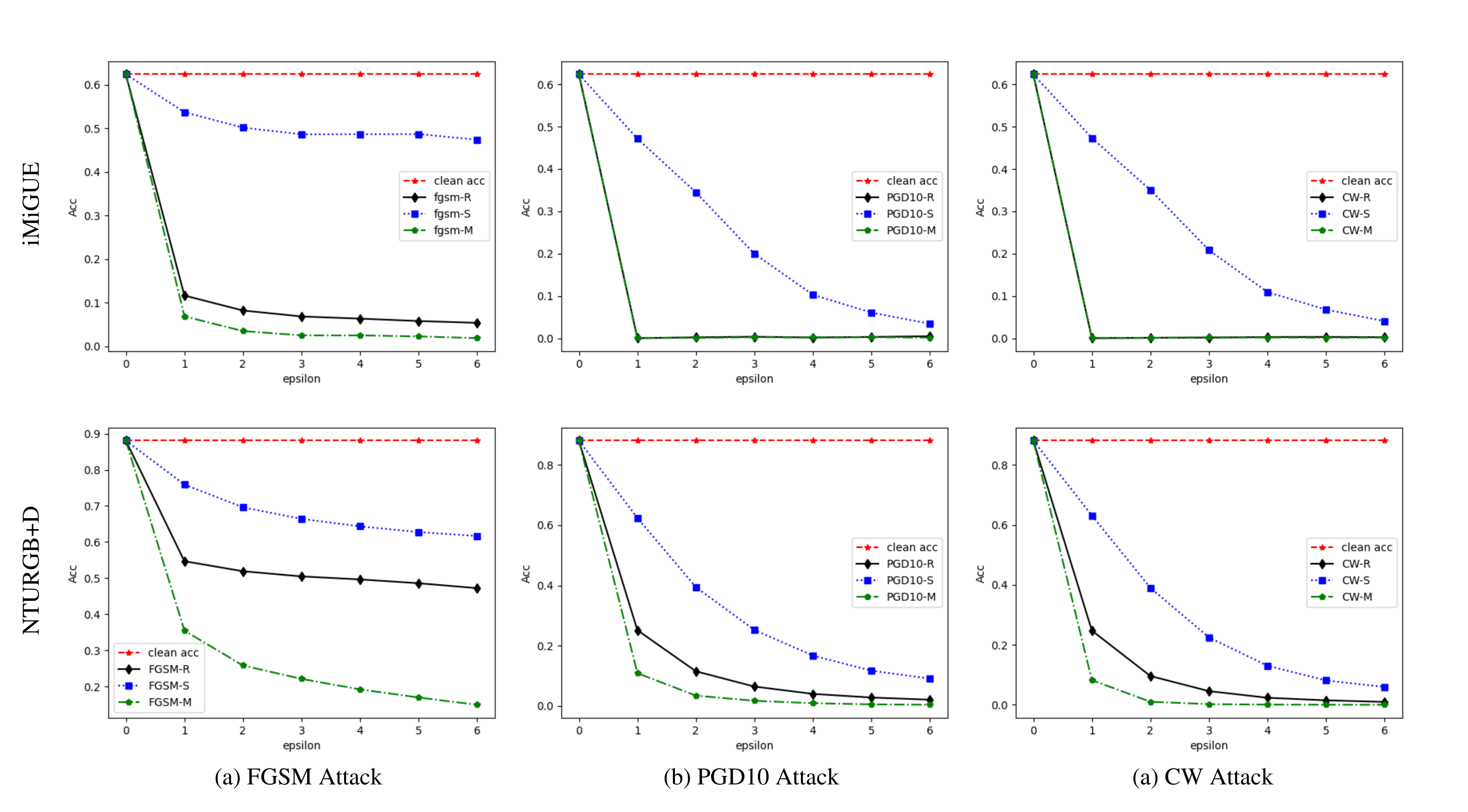}
    \caption{Accuracy against three types of adversarial
attacks based on the iMiGUE dataset \cite{Liu2021iMiGUEAI} and the NTU-RGB+D dataset \cite{Shahroudy2016NTURA}, namely adversarial robustness. The x-axis indicates the attack strength $\varepsilon$ 
($\times \frac{2}{255} $). FGSM-S, and FGSM-M respectively denote the FGSM attack on the RGB modality, skeleton modality, and multiple modalities simultaneously, and so on. 'Clean acc' represents the accuracy after multimodal fusion on clean data.The findings manifest that, as the intensity of attacks increases, the robustness of the skeleton modality exhibits a comparatively gradual and smooth decline in contrast to the sharp decrease observed in both the RGB modality and multimodal fusion. Therefore, we confirm the assertion that the skeleton modality shows higher robustness than the RGB modality.}
    \label{fig:image}
\end{figure}

To understand the robustness of various modalities, we analyzed the RGB and skeleton modalities under different attacks and intensities. \cref{fig:image} shows our experiments on the NTU-RGB+D \cite{Shahroudy2016NTURA} and iMiGUE \cite{Liu2021iMiGUEAI} datasets, evaluating their response to three types of attacks. As attack intensity increases, the skeletal modality's accuracy declines slower and smoother than the RGB and multimodal modalities, which drop sharply. This indicates that slight perturbations cause significant errors in the RGB modality, highlighting the skeletal modality's higher robustness. Both datasets show the same trend. This aligns with findings by Yan \etal \cite{2018Spatial}, who noted that bone and joint trajectories are robust to lighting and scene variations. Skeleton data lacks intricate information like lighting and background details, making it easier to capture human actions. Multimodal fusion results in the lowest accuracy, as found in \cite{Tian2021CanAI}, suggesting it does not enhance and may even diminish robustness.

Based on this observation, we propose an Attention-based Modality Reweighter (AMR) to enhance the robustness of RGB-skeleton action recognition models, shown in \cref{fig:framework}. Our backbone consists of two branch networks, I3D \cite{2017Quo} and HCN \cite{2018Co}, extracting features from RGB videos and skeleton, respectively. Although I3D and HCN are not SOTA, their simple design suits our approach. Our goal is to enhance model robustness, not just achieve the highest action recognition rate. Using a simple model eliminates unnecessary factors. In the final blocks, we incorporate AMR to connect a two-stream network. AMR has two attention layers assigning weights to features from both modalities, helping the model learn robust features, especially against adversarial attacks. After reweighted features pass through global pooling layers (GAP if necessary), two classification vectors are derived, concatenated, and fed into a fully connected layer (FC) for auxiliary classification results. AMR generates three outputs: reweighted features of both modalities and auxiliary classification predictions. The adjusted features continue forward propagation, while auxiliary classification results are used in the loss function. Further details are in \cref{sec:3}.

In summary, our contributions are:
\begin{itemize}
  \item[$\bullet$] We conducted empirical analysis comparing the robustness of different modalities, showing the superior robustness of the skeleton modality over RGB.
  \item[$\bullet$] We propose a novel module, Attention-based Modality Reweighter (AMR), with a new loss function. It supports both adversarial and standard training, reweighting modalities to enhance robustness.
  \item[$\bullet$] We conducted comprehensive experiments on popular benchmark datasets, evaluating performance against classical adversarial attacks. The results show that our method enhances the robustness of multimodal action recognition models, addresses robustness discrepancies across modalities, and achieves new SOTA performance without additional data.
\end{itemize}

\section{Related Work}

\subsection{RGB-Skeleton Action Recognition}

Conventional approaches to action recognition primarily rely on RGB video as the main input modality. Yan \etal \cite{2018Spatial} first introduced deep learning into skeleton-based action recognition. RGB videos provide comprehensive visual information about actions, including human attributes, attire, colors, spatial relationships, and contextual background details. Skeletal data focuses on pose information, offering insights into human posture and alignment through skeletal key-points. RGB and skeletal modalities provide synchronized and complementary information. Their integration enhances action recognition by offering more comprehensive and accurate information. Numerous studies in RGB-skeleton action recognition have emerged \cite{Duan2021RevisitingSA,Yu2022MMNetAM,Joze2019MMTMMT}. While clean data fusion aids action recognition, the impact on robustness when RGB-skeleton modalities face attacks is unclear. This paper uses RGB-skeleton action recognition to investigate the robustness of multimodal learning against adversarial attacks.

\subsection{Adversarial Attack}
Since \cite{Goodfellow2014ExplainingAH} finds the adversarial perturbations in neural networks, generating adversarial images to attack deep networks have attracted great interests. Lots of attack methods have been proposed to craft adversarial examples which can fool deep neural networks. The most famous attack is Fast Gradient Sign Method (FGSM) \cite{Goodfellow2014ExplainingAH}, which only needs one step to create adversarial examples. Further, Madry \etal \cite{2017Towards} proposed the Project Gradient Descent (PGD) attack, which is the most used adversarial attack in the adversarial research field. Boundary-based attacks such as deepfool \cite{2015DeepFool} and CW \cite{2016Towards} also make the model more challenging. Duan \etal \cite{duan2022novel} proposed a multi-sample generation model for black-box model attacks, called MsGM. AutoAttack (AA) is an ensemble attack, which consists of three white-box attacks (APGD-CE \cite{Croce2020ReliableEO}, APGD-DLR \cite{Croce2020ReliableEO}, and FAB \cite{Croce2019MinimallyDA}) and one black-box attack (Square attack \cite{Andriushchenko2019SquareAA}) for a total of four attack methods.Xu \etal \cite{xu2023a2sc} investigate the undiscovered adversarial attacks in the unsupervised clustering domain, a field of equal significance, and propose the concept of the adversarial set and adversarial set attack for clustering.

\subsection{Adversarial Defense}
With the development of adversarial attacks, there is also a lot of work to improve the adversarial robustness of deep neural networks \cite{He2015DeepRL}. Adversarial training \cite{2017Towards} is the most effective way to defend against adversarial examples and serves as our fundamental training approach. TRADES \cite{Zhang2019TheoreticallyPT} analysis the trade-off between clean accuracy and adversarial accuracy, and uses the Kullback-Leibler divergence \cite{Kullback1951OnIA} to balance them. Misclassification Aware adveRsarial Training (MART) \cite{Wang2020ImprovingAR} emphasizes the misclassified examples. Adversarial logit pairing (ALP) \cite{Kannan2018AdversarialLP} and its variant Adaptive Adversarial Logits Pairing (AALP) \cite{wu2023adaptive} define pairing loss that pulls adversarial logit and natural logit together. Adversarial Neural Pruning (ANP) \cite{2019Adversarial} uses Bayesian methods to prune vulnerable features. Zhang \etal \cite{2020Geometry} uses sample reweighting techniques to improve the adversarial robustness against PGD attack.

All the above works are in the fields of image and video processing. Tian \etal \cite{Tian2021CanAI} is most relevant to our research, investigating audio-visual model robustness under multimodal attacks. Their experiments show that integrating audio and visual components under multimodal attacks does not enhance but decreases model robustness. This indicates that audio-visual models are vulnerable to multimodal adversarial attacks and that integration may decrease robustness. To mitigate these attacks, they propose an multimodal defense approach based on an audio-visual dissimilarity constraint (MinSim) and External Feature Memory banks (ExFMem), representing the SOTA in multimodal defense. However, ExFMem's storage of external features requires additional disk space and does not consider variations in robustness across different modalities. Our proposed method addresses these challenges.

\section{Methodology}\label{sec:3}
In this section, we first introduce the Standard Adversarial Training (SAT)~\cite{2017Towards}. Then, we will describe the Attention-based Modality Reweighter (AMR) in detail.

\subsection{Standard Adversarial Training}\label{sec:3.1}
Given a standard training data set $D=\{(x_{i},y_{i})\}_{i=1}^{n}$ with n
examples and $C$ classes, where $x_{i}\in\mathbb{R}^{d}$ is the natural example and $y_{i}\in\{0,1...,C-1\}$ is corresponding ground-truth label. Standard training (ST) achieves good classification performance by minimizing the empirical risk on the training dataset, formulated as \cref{eq:ST}.
\begin{equation}
  \operatorname*{min}_{\theta\in\Theta}{\frac{1}{n}}\sum_{i=1}^{n}l(f_{\theta}(x_{i}),y_{i}),
  \label{eq:ST}
\end{equation}
where $l$ is the cross-entropy loss widely used in classification tasks, $f\colon\mathbb{R}^{d}\to\mathbb{R}^{C}$ is the neural network parameterized by $\theta$ and $\Theta$ is the parameter space of $\theta$. The neural network trained in this manner, however, does not exhibit satisfactory prediction accuracy when faced with adversarial examples. To solve this problem, Madry \etal \cite{2017Towards} used adversarial data to train the neural network, which can be formulated as a min-max optimization problem:
\begin{equation}
    \operatorname*{min}_{\theta\in\Theta}{\frac{1}{n}}\sum_{i=1}^{n}I(f_{\theta}(x_{i}^{\prime}),y_{i}),
    \label{eq:SAT}
\end{equation}
where
\begin{equation}
    x_i^\prime  = \mathop {\arg \max }\limits_{{x^\prime } \in {{\mathcal B}_\varepsilon }[{x_i}]} l({f_\theta }({x^\prime }),{y_i}).
    \label{eq:advx}
\end{equation}
${\mathcal{B}}_{\varepsilon}[x_{i}]$ is the sampling space of the adversarial example $x_i^\prime$, which is bounded in the $L_{p}$-norm neighbor space of the natural example $x_{i}$, \emph{i.e.}, ${{\mathcal{B}}_{\varepsilon }}[{{x}_{i}}]=\{{{x}^{\prime }}\in {{\mathbb{R}}^{d}}|{{\left\| {{x}^{\prime }}-{{x}_{i}} \right\|}_{p}}\le \varepsilon \}$, and is the closed ball of radius $\varepsilon >0$ centered at $x_{i}$.$\delta =x_{i}^{\prime }-{{x}_{i}}$ is the adversarial perturbation. In this paper, only $p=\infty$ is considered. It implies the optimization of adversarialy robust network, with one step maximizing loss to find adversarial data and one step minimizing loss on the adversarial data \emph{w.r.t.} the network parameters $\theta$. To generate adversarial data, SAT \cite{2017Towards} uses PGD to approximately solve the inner maximization of \cref{eq:advx}. Namely, given a starting point $x_{0}^{\prime }\in {{\mathbb{R}}^{d}}$ and step size $\alpha > 0$, PGD works as follows:
\begin{equation}
    x_{t+1}^{\prime }=\prod\limits_{{{\mathcal{B}}_{\varepsilon }}[{{x}_{0}}^{\prime }]}{(x_{t}^{\prime }+\alpha sign({{\nabla }_{x_{t}^{\prime }}}l({{f}_{\theta }}(x_{t}^{\prime }),y))}),\forall t\ge 0
    \label{eq:PGD}
\end{equation}
until a certain stopping criterion is satisfied. For example, the criterion can be a fixed number of iterations K, namely the PGD-K algorithm \cite{2017Towards}. In \cref{eq:PGD}, $l$ is the loss function in \cref{eq:advx}; $x_{0}^{\prime }$ refers to natural data or natural data corrupted by a small Gaussian
or uniform random noise; $y$ is the corresponding label for natural data; $x_{t}^{\prime }$ is adversarial data at step $t$; and $\prod\limits_{{{\mathcal{B}}_{\varepsilon }}[{{x}_{0}}^{\prime }]}{(\cdot )}$ is the projection function that projects the adversarial data back into the $\epsilon$-ball centered at $x_{0}^{\prime }$ if necessary.

\subsection{Multimodal Adversarial Training}
Let $x_{R,i}$ be an input RGB video frame, $x_{s,i}$ be an input skeleton data, and $y_{i}$ be the corresponding groundtruth label for the multisensory input: $\{x_{R,i}^{{}},x_{s,i}^{{}}\}$. Since there are multiple inputs, we can divide our multimodal attack into two categories: unimodality attacks that only generate RGB adversarial example $x_{R,i}^{\prime}$ a or skeleton adversarial example $x_{s,i}^{\prime}$, and RGB-skeleton multimodal attacks that generate both RGB and skeleton adversarial examples: $\{x_{R,i}^{\prime },x_{s,i}^{\prime }\}$. We can extend the min-max formulation \cref{eq:SAT} and \cref{eq:advx} to multimodal data \cite{Tian2021CanAI}, formulated as follows:
\begin{equation}
    \underset{\theta \in \Theta }{\mathop{\min }}\,\frac{1}{n}\sum\limits_{i=1}^{n}{l({{f}_{\theta }}(x_{R,i}^{\prime },x_{s,i}^{\prime }),{{y}_{i}})},
    \label{eq:multiAT}
\end{equation}
where
\begin{equation}
    \{x_{R,i}^{\prime },x_{s,i}^{\prime }\}=\underset{\begin{smallmatrix} 
 {{x}_{R}}^{\prime }\in {{\mathcal{B}}_{{{\varepsilon }_{R}}}}[{{x}_{R,i}}], \\ 
 {{x}_{s}}^{\prime }\in {{\mathcal{B}}_{{{\varepsilon }_{s}}}}[{{x}_{s,i}}] 
\end{smallmatrix}}{\mathop{\arg \max }}\,l(f(x_{R}^{\prime },x_{s}^{\prime }),{{y}_{i}}),
    \label{eq:multiadvx}
\end{equation}
where ${{\mathcal{B}}_{{{\varepsilon }_{\text{R}}}}}[{{x}_{R,i}}]=\{x_{R}^{\prime }\in {{\mathbb{R}}^{{{d}_{R}}}}|{{\left\| x_{R}^{\prime }-{{x}_{R,i}} \right\|}_{p}}\le {{\varepsilon }_{R}}\}$ , ${{\mathcal{B}}_{{{\varepsilon }_{s}}}}[{{x}_{s,i}}]=\{x_{s}^{\prime }\in {{\mathbb{R}}^{{{d}_{s}}}}|{{\left\| x_{s}^{\prime }-{{x}_{s,i}} \right\|}_{p}}\le {{\varepsilon }_{s}}\}$. ${{\delta }_{R}}=x_{R,i}^{\prime }-{{x}_{R,i}}$ and ${{\delta }_{s}}=x_{s,i}^{\prime }-{{x}_{s,i}}$ are video adversarial perturbation and skeleton adversarial perturbation, respectively. The symbols used in the formulas presented in this section retain the same meanings as those discussed in \cref{sec:3.1}. With the adversarial objective, the attacker will maximize the loss function by seeking small perturbations within allowed budgets, and try to push the trained model to make incorrect predictions. For unimodality attacks, either $\delta_{R}$ or $\delta_{s}$ is 0. Likewise, \cref{eq:PGD} can be extended to generate multimodal adversarial samples:
\begin{equation}
\left\{ 
\begin{aligned}
& x_{R,t+1}^{\prime }=\prod\limits_{{{\mathcal{B}}_{{{\varepsilon }_{R}}}}[{{x}_{R,0}}^{\prime }]}{(x_{R,t}^{\prime }+\alpha sign({{\nabla }_{x_{R,t}^{\prime }}}l({{f}_{\theta }}(x_{R,t}^{\prime },x_{s,t}^{\prime }),y))}), \\ 
& x_{s,t+1}^{\prime }=\prod\limits_{{{\mathcal{B}}_{{{\varepsilon }_{s}}}}[{{x}_{s,0}}^{\prime }]}{(x_{s,t}^{\prime }+\alpha sign({{\nabla }_{x_{s,t}^{\prime }}}l({{f}_{\theta }}(x_{R,t}^{\prime },x_{s,t}^{\prime }),y))}),
\end{aligned}
\label{eq:PGDmulti}
\right.
\end{equation}
where, $\forall t\ge 0$.

\subsection{Attention-based Modality Reweighter}\label{sec:3.3}

Reviewing \cref{fig:image}, numerous experimental results show that the robustness of the skeleton and RGB modalities differs: the skeleton modality is significantly more robust. As attack intensity increases, the skeleton modality's accuracy declines gradually and smoothly, while the RGB and multimodal approaches drop sharply. These results align with the conclusions in \cite{2018Spatial,Tian2021CanAI,yuan2024auformer}. Motivated by this finding, we aim to enable the multimodal model to assign appropriate weights to the features of the two modalities, making the model learn more robust features against adversarial attacks. Specifically, more robust features should receive greater weight, while less robust features should get smaller weights.

Our proposed AMR is illustrated in \cref{fig:module}. Specifically, given a mini-batch of multi-modal natural sample $\{{{x}_{R}},{{x}_{s}}\}$ ($B$ in total), we can obtain corresponding adversarial $\{x_{R}^{\prime },x_{s}^{\prime }\}$ samples by using algorithms such as PGD, as mentioned in \cref{eq:multiadvx}. Then, we extract latent features for both the video and skeleton adversarial samples from the hidden layers of the two branch networks, denoted as $X_{R}^{\prime }\in {{\mathbb{R}}^{B\times {{C}_{R}}\times {{T}_{R}}\times H\times W}}$ and $X_{s}^{\prime }\in {{\mathbb{R}}^{B\times {{C}_{s}}\times {{T}_{s}}\times V}}$ respectively, where $B$ is the input batch size, $C_{R}$ and $C_{s}$ are respectively the number of output channels for the video and skeleton, $T_{R}$ and $T_{s}$ are the number of frames of the video sequence and the skeleton sequence, respectively. $H$ and $W$ are the spatial dimensions of the video. $V$ is the number of vertices in each frame of skeleton data. Subsequently, we create two attention weight matrices, ${{W}_{R}}\in {{\mathbb{R}}^{1\times {{C}_{R}}\times {{T}_{R}}\times H\times W}}$ and ${{W}_{s}}\in {{\mathbb{R}}^{1\times {{C}_{s}}\times {{T}_{s}}\times V}}$, based on the size of the features, noting that these matrices will be updated alongside the other parameters of the network. After replicating the two attention weight matrices $B$ times along the batch size dimension, they are respectively multiplied with the corresponding extracted latent features, formulated as follows:
\begin{figure}
    \centering
    \includegraphics[width=1\linewidth]{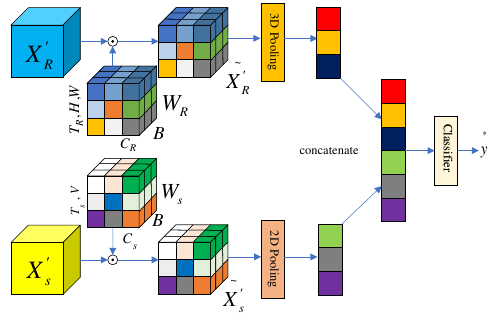}
    \caption{Architecture of AMR for two modalities. $X_{R}^{\prime}$ and $X_{s}^{\prime}$, that represent the features at a given layer of two unimodal network, are the inputs to the module. For better visualization, we represent the spatiotemporal dimensions on a single axis.}
    \label{fig:module}
\end{figure}
\begin{equation}
\left\{
\begin{aligned}
  \overset{\sim}{\mathop{X_{R}^{\prime }}} &= X_{R}^{\prime} \otimes W_{R}, \\ 
  \overset{\sim}{\mathop{X_{s}^{\prime }}} &= X_{s}^{\prime} \otimes W_{s},
  \label{eq:matric}
\end{aligned}
\right.
\end{equation}
where $\otimes$ represents element-wise multiplication. At present, $\overset{\sim}{\mathop{X_{R}^{\prime }}}\,\in {{\mathbb{R}}^{B\times {{C}_{R}}\times {{T}_{R}}\times H\times W}}$ and $\overset{\sim}{\mathop{X_{s}^{\prime }}}\,\in {{\mathbb{R}}^{B\times {{C}_{s}}\times {{T}_{s}}\times V}}$ embody features adjusted by the attention weight matrices. Consecutively, they are separately fed into a 3D pooling layer and a 2D pooling layer, and after flattening, the results are as follows:
\begin{equation}
\left\{ 
\begin{aligned}
{{Z}_{R}} &=\frac{1}{{{T}_{R}}\times H\times W}\sum\limits_{t=1}^{{{T}_{R}}}{\sum\limits_{h=1}^{H}{\sum\limits_{w=1}^{W}{\overset{\sim}{\mathop{X_{R}^{\prime }}}\,}}}, \\ 
{{Z}_{s}} &=\frac{1}{{{T}_{s}}\times V}\sum\limits_{t=1}^{{{T}_{s}}}{\sum\limits_{v=1}^{V}{\overset{\sim}{\mathop{X_{s}^{\prime }}}\,}},
\label{eq:flatten}
\end{aligned} 
\right.
\end{equation}
where
\begin{equation}
\left\{ 
\begin{aligned}
& {{Z}_{R}}\in {{\mathbb{R}}^{B\times {{C}_{R}}}}, \\ 
& {{Z}_{s}}\in {{\mathbb{R}}^{B\times {{C}_{s}}}}, \\ 
\end{aligned}
\right.
\end{equation}

Afterward, concatenate ${{Z}_{R}}$ and ${{Z}_{s}}$ along their respective channel dimensions to obtain ${{Z}_{mul}}\in {{\mathbb{R}}^{B\times ({{C}_{R}}+{{C}_{s}})}}$, which is a vector that can be leveraged for classification. Finally, we can feed ${{Z}_{mul}}$ into a fully connected layer (FC) to generate auxiliary classification prediction $\overset{*}{\mathop{y}}\,\in {{\mathbb{R}}^{B\times N}}$. The above is a detailed description of the structure of AMR. In this module, three outputs will be provided: the features $\{ \overset{\sim}{\mathop{X_{R}^{\prime }}}\,,\overset{\sim}{\mathop{X_{s}^{\prime }}}\, \}$ of the two modalities after attention weight adjustment, as well as the auxiliary classification prediction $\overset{*}{\mathop{y}}$. The reweighted features continue to undergo forward propagation, with $\overset{*}{\mathop{y}}$ being utilized for the computation of the loss function.

We can integrate $S$ AMRs into distinct hidden layers of a multimodal network. AMR can be deemed as an auxiliary component of the network, amenable to standard training or adversarial training in conjunction with the backbone network. We devise a novel loss function. Taking S instances of AMR inserted into the network as an example, the AMR loss function can be defined as follows:
\begin{equation}
    {{L}_{AMR}}=l({{y}^{\prime }},y)+\frac{1}{S}\sum\limits_{i=1}^{S}{l({{\overset{*}{\mathop{y}}\,}_{i}},y)},
    \label{eq:AMR}
\end{equation}
where ${{y}^{\prime }}={{f}_{\theta }}(x_{R}^{\prime },x_{s}^{\prime })$ signifies the ultimate predictive output for adversarial samples. $y$ is the corresponding groundtruth label. $l(\cdot )$ denotes the cross-entropy loss function. \cref{eq:AMR} encompasses the auxiliary classification results from multiple hidden layers, serving the purpose of deep supervision. It compels the network to learn more robust features at intermediate layers. To simultaneously consider both clean accuracy and robust accuracy, we also incorporate the outputs of natural samples into the loss function. The overall objective function for adversarial training of our AMR strategy is as follows:
\begin{equation}
    {{L}_{total}}=l(\overset{\hat{\ }}{\mathop{y}}\,,y)+\lambda {{L}_{AMR}},
    \label{eq:total}
\end{equation}
where $\overset{\hat{\ }}{\mathop{y}}\,={{f}_{\theta }}({{x}_{R}},{{x}_{s}})$ denotes the prediction output for nature data, and $\lambda$ functions as a hyperparameter to modulate the balance between the two loss components, allowing for customization to accommodate diverse datasets.

\section{Experiment}

\subsection{Dataset}
\textbf{NTU-RGB+D}~\cite{Shahroudy2016NTURA} is a well-known large-scale multimodal human action recognition dataset. It contains 56,880 samples captured from 40 subjects performing 60 classes of activities at 80 view-points. Each action clip includes up to two people on the RGB video as well as 25 body joints on 3D coordinate space. The dataset is split in two ways: Cross-subject (X-Sub) and Cross-view (X-View), for which action subjects, camera views are different in training and validation. Unless otherwise specified, we conduct experiments on the X-sub splits for NTU-RGB+D.

\textbf{iMiGUE}~\cite{Liu2021iMiGUEAI} is a new dataset for the emotional artificial intelligence research: identity-free video dataset for Micro-Gesture Understanding and Emotion analysis (iMiGUE). iMiGUE comprises 36 female and 36 male players whose ages are between 17 and 38. After processing, a total of 18,419 video clips were included, divided into 32 action categories. The corresponding human skeleton data is also available.

\subsection{Experimental Setup}\label{sec:4.2}
\textbf{Training setup:} The overall architecture utilized in the experiments is illustrated in Figure 1. For iMiGUE and NTU-RGB+D datasets, we respectively insert 1 and 3 AMRs forward from the last fully connected (FC) layer of both the I3D and HCN branch networks. We employe SGD (momentum 0.9, batch size 16) to train the model for 40 epochs on the iMiGUE dataset and 80 epochs on the NTU-RGB+D dataset with weight decay 5e-4 and initial learning rate 0.01. To adjust the learning rate adaptively, we utilize the cosine annealing learning rate decay strategy, where $T_{max}$ corresponds to the respective number of epochs. For the internal maximization process, we use PGD10 attack to simultaneously obtain adversarial samples for two modalities, with a random start, step size $2.0/255$, and perturbation size $\epsilon_{R},\epsilon_{s} = 8.0/255$. We use $\lambda = 1$ in all experiments.

\textbf{Evaluation setup:} We report the clean accuracy on natural examples and the adversarial accuracy on adversarial examples. We follow the widely used protocols in the adversarial research field. We consider three popular attack methods: FGSM \cite{Goodfellow2014ExplainingAH}, PGD20 \cite{2017Towards}, and CW \cite{2016Towards} (optimized by PGD30).

\subsection{Model Robustness under Multimodal Attacks}\label{sec:4.3}
\begin{table}[]
\begin{center}
\resizebox{\columnwidth}{!}{
\begin{tabular}{c|c|c|cccc|c|c}
\hline
Dataset                   & Attack & \ding{51}RS              & \ding{55}R     & \ding{55}S     & \ding{55}RS    & Avg   & Unimodal \ding{51}R                  & Unimodal \ding{51}S                  \\ \hline
\multirow{3}{*}{NTURGB+D} & FGSM   & \multirow{3}{*}{88.09} & 50.48 & 66.42 & 22.12 & 46.34 & \multirow{3}{*}{81.46} & \multirow{3}{*}{81.46} \\ 
                          & PGD10  &                        & 6.39  & 25.18 & 1.66  & 11.08 &                        &                        \\ 
                          & CW     &                        & 4.55  & 22.46 & 0.21  & 9.03  &                        &                        \\ \hline
\multirow{3}{*}{iMiGUE}   & FGSM   & \multirow{3}{*}{62.37} & 6.86  & 48.61 & 2.52  & 19.03 & \multirow{3}{*}{59.48} & \multirow{3}{*}{46.86} \\
                          & PGD10  &                        & 0.35  & 19.94 & 0.24  & 6.84  &                        &                        \\ 
                          & CW     &                        & 0.20  & 20.82 & 0.26  & 7.09  &                        &                        \\ \hline
\end{tabular}}
\label{tab:1}
\end{center}
\caption{Various accuracy(\%) on NTU-RGB+D and iMiGUE datasets under different attack methods. \ding{55}R \ding{55}S and \ding{55}RS denote that only RGB, only skeleton, and both RGB video and skeleton inputs are attacked, respectively. We set $\epsilon_{R}$ and $\epsilon_{s}$ as 6/255 respectively for \ding{55}R \ding{55}S and \ding{55}RS. The symbol: \ding{51} means that inputs are clean. The baselines: Unimodal \ding{51}R and Unimodal \ding{51}S models are are their respective unimodality branch models.}
\end{table}

In \cref{fig:image}, we qualitatively observed the variation trends of model robustness under different attack scenarios. Here, we quantitatively analyze the diverse facets of model robustness under multimodal adversarial attacks. \cref{tab:1} presents various aspects of the model's robustness on the NTU-RGB+D and iMiGUE datasets under different attack scenarios. To better interpret the multimodal robustness, we also include results from two baselines: Unimodal R and Unimodal S, which are two unimodality models and only use video and skeleton modalities, respectively. Clearly, all of the three attack methods can significantly decrease recognition results. The results demonstrate that at the same level of attack intensity, the accuracy of the skeleton modality is notably higher than that of the RGB modality and the multimodal, indicating the stronger resilience of the skeleton modality to interference. When we leverage different attack strategies to perform unimodality attacks on the NTU-RGB+D and iMiGUE datasets, RGB-skeleton models: \ding{55}R and \ding{55}S are always inferior to Unimodal \ding{51}S and Unimodal \ding{51}R, respectively. Note that \ding{55}R and \ding{55}S have clean skeleton and RGB modalities, respectively. This discovery suggests that when one modality input is subjected to an attack, the multimodal integration weakens the recognition performance, consistent with the conclusion in \cite{Tian2021CanAI}.

\begin{table}[]
\begin{center}
\resizebox{\columnwidth}{!}{
\begin{tabular}{c|c|cccc|c}
\hline
Defence(NTURGB+D) & \ding{51}RS             & \ding{55}R              & \ding{55}S              & \ding{55}RS             & Avg            & RI             \\ \hline
None              & \textbf{88.09}          & 2.09           & 11.84          & 0.15           & 4.69           & 0              \\
AT\cite{2017Towards} with \ding{55}R        & 75.59          & \textbf{75.59} & 4.75           & 4.74           & 28.36          & 11.17          \\
AT\cite{2017Towards} with \ding{55}S        & 78.44          & 2.09           & 78.31          & 2.06           & 27.49          & 13.15          \\
AT\cite{2017Towards} with \ding{55}RS       & 61.42          & 61.41          & 40.41          & 40.32          & 47.38          & 16.02          \\
MinSim\cite{Tian2021CanAI}            & 80.85          & 2.93           & 80.85          & 3.03           & 28.85          & 16.92          \\
ExFMem\cite{Tian2021CanAI}            & 84.19 & 0.00           & 60.04          & 0.00           & 20.01          & 11.42          \\
MinSim+ExFMem\cite{Tian2021CanAI}     & 82.81          & 2.02           & 75.77          & 2.14           & 26.64          & 16.67          \\
AMR(Ours)      & 81.65          & 68.54          & \textbf{80.40} & \textbf{66.52} & \textbf{71.82} & \textbf{60.69} \\ \hline
Defence(iMiGUE)   & \ding{51}RS             & \ding{55}R              & \ding{55}S              & \ding{55}RS             & Avg            & RI             \\ \hline
None              & \textbf{62.37} & 0.11           & 6.97           & 0.07           & 2.38           & 0              \\
AT\cite{2017Towards} with \ding{55}R        & 47.32          & \textbf{47.32} & 0.00           & 0.00           & 15.77          & -1.66          \\
AT\cite{2017Towards} with \ding{55}S        & 60.20          & 0.00           & \textbf{59.87} & 0.00           & 19.96          & 15.41          \\
AT\cite{2017Towards} with \ding{55}RS       & 39.01          & 39.01          & 31.89          & 31.95          & 34.28          & 8.54           \\
MinSim\cite{Tian2021CanAI}            & 60.60          & 0.00           & 53.28          & 0.00           & 17.76          & 13.61          \\
ExFMem\cite{Tian2021CanAI}            & 60.53          & 0.00           & 43.00          & 0.00           & 14.33          & 10.11          \\
MinSim+ExFMem\cite{Tian2021CanAI}     & 60.62          & 0.00           & 55.05          & 0.00           & 18.35          & 14.22          \\
AMR(Ours)      & 43.89          & 43.89          & 32.85          & \textbf{32.94} & \textbf{36.56} & \textbf{15.70} \\ \hline
\end{tabular}}
\end{center}
\caption{The various accuracy(\%) metrics on the NTU-RGB+D and iMiGUE datasets are presented with different defense methods. Here, we utilize PGD20 (perturbation budgets $\epsilon_{R},\epsilon_{s} = 8/255$,step size $\alpha = 2/255$) to generate adversarial samples for assessing robustness accuracy. The best results among the metrics are denoted in bold.}
\label{tab:2}
\end{table}

\subsection{Robustness Analysis and Evaluation}
To validate the effectiveness of the proposed AMR, we compared it with several baselines and SOTA methods. For the sake of fairness, the SOTA methods we compared to are also multimodal defense strategies, rather than unimodal defense approaches.

\textbf{Baselines and SOTAs:} 1) None: RGB-skeleton multimodal network without any defense; 2) AT \cite{2017Towards} with \ding{55}R: adversarial training using only adversarial samples from the RGB video modality; 3) AT \cite{2017Towards} with \ding{55}S: adversarial training using only adversarial samples from the skeleton modality; 4) AT \cite{2017Towards} with \ding{55}RS: adversarial training using adversarial samples from both the RGB video and the skeleton modality simultaneously; 5) MinSim \cite{Tian2021CanAI}: a recent SOTA multimodal defense mechanism, namely the dissimilarity constraint to encourage multimodal dispersion and unimodal compactness; 6) ExFMem \cite{Tian2021CanAI}: a recent SOTA multimodal defense mechanism,namely adopting external feature memory banks to denoise attacked video and skeleton examples at a feature level; 7) MinSim+ExFMem \cite{Tian2021CanAI}: the combination of methods MinSim and ExFMem; 8) AMR: our proposed defence method (certainly based on AT with \ding{55}RS)

\textbf{Evaluation Metrics:} To evaluate the performance of different defense methods and facilitate a fair comparison, we follow the protocol proposed in \cite{Tian2021CanAI}. We use recognition accuracy as the
metric. Results from both the nature samples: \ding{51}RS and adversarial samples: \ding{55}R, \ding{55}S, and \ding{55}RS are computed. Since there are multiple defense results under different multimodal attacks for a single method, we also use the averaged accuracy:
\begin{equation}
    Avg = \frac{1}{3}(\text{\ding{55}}R + \text{\ding{55}}S + \text{\ding{55}}RS),
    \label{eq:avg}
\end{equation}
as an overall metric to evaluate robustness of different defenses. To comprehensively assess the model's clean accuracy and robust accuracy, Tian \etal \cite{Tian2021CanAI} propose a relative improvement (RI) metric:
\begin{equation}
    RI = (\text{\ding{51}}RS_{m} + Avg_{m}) - (\text{\ding{51}}RS_{n} + Avg_{n}),
    \label{eq:RI}
\end{equation}
where results from both clean samples and adversarial samples are considered, and the $m$ refers to a defense method and $n$ refers to a base model, which is the baseline: None in our experiments. If a defense method decreases clean data performance, the RI will penalize it accordingly.

\textbf{Result Analysis:} \cref{tab:2} and display various accuracy metrics against PGD20 attack on the NTU-RGB+D and iMiGUE datasets with different defense methods. See more results in \cref{tab:FGSM attack} and \cref{tab:3} in the Supplementary. The \ding{51}RS, representing the accuracy on clean data, is highest with the None, \emph{i.e.}, the baseline. This aligns with adversarial training and robustness theory, as incorporating adversarial examples during training is bound to compromise clean accuracy. AT enhances robustness that None lacks. It can be observed that training with adversarial samples from different modalities significantly enhances the robustness of the corresponding modality. AT with \ding{55}R and AT with \ding{55}S achieve the highest values on \ding{55}R and \ding{55}S, respectively. However, they exhibit almost no robustness on the modalities without the use of adversarial samples. AT with \ding{55}RS strikes a balance between the robustness of various modalities. The most relevant method to our work, MinSim/ExFMem, achieves high performance in \ding{51}RS and \ding{55}S, but its accuracy is very poor in \ding{55}R and \ding{55}RS. This is because this method does not consider the differences in the inherent robustness of different modalities and does not specially handle the modalities that are more susceptible to attacks. From the results, it is evident that our AMR outperforms all compared methods in terms of \ding{55}RS and Avg (representing robust accuracy) as well as the comprehensive performance indicator RI. AMR visibly boosts the robustness of the vulnerable RGB modality by taking into account the characteristics of each modality.

\begin{figure*}
\begin{center}
\includegraphics[width=0.87\linewidth]{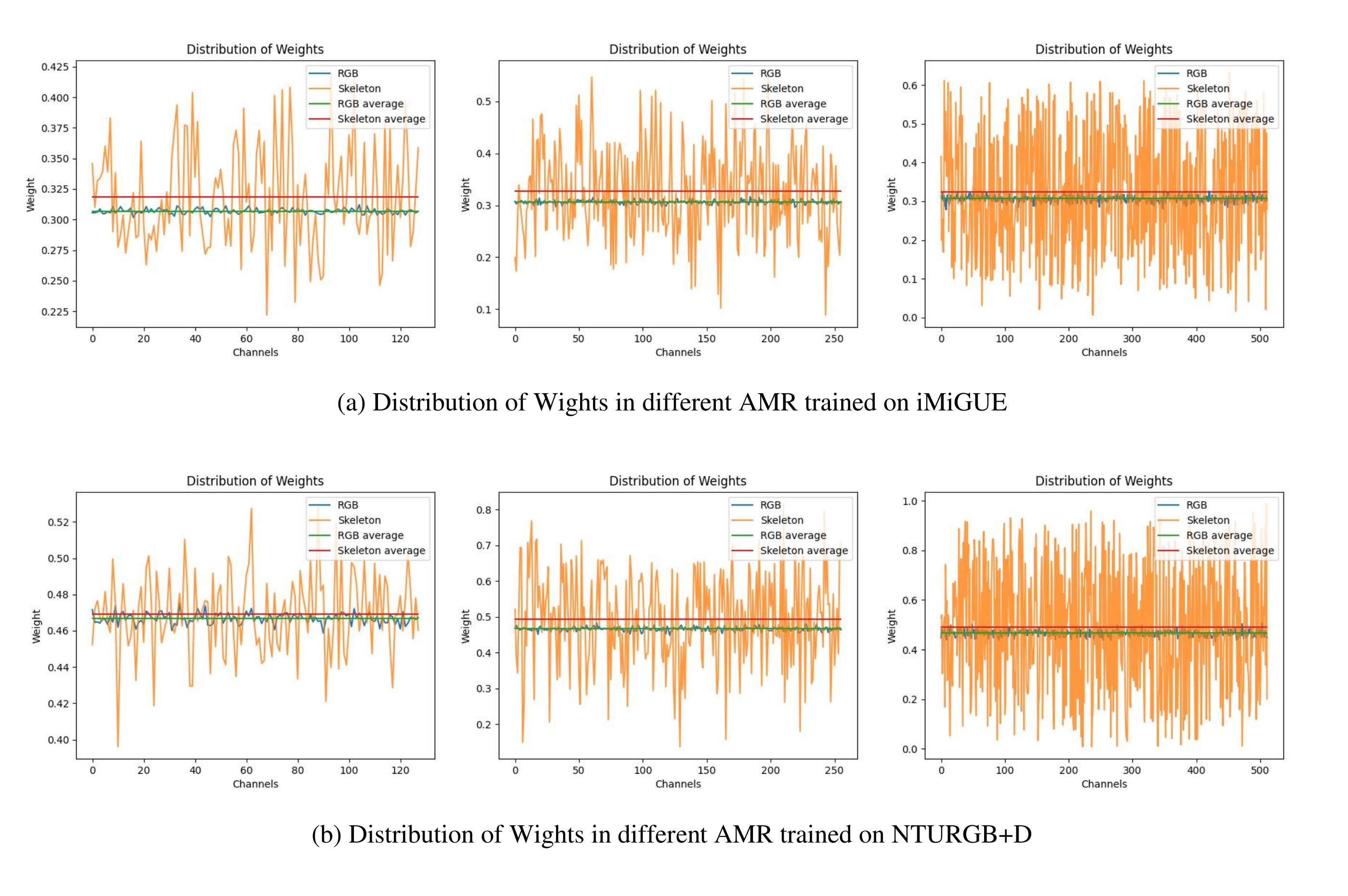}
\end{center}
    \caption{The distribution of weight matrices in different AMRs trained on two datasets. The $x$-axis represents the channels, and the $y$-axis represents the average weight values corresponding to each channel.}
    \label{fig:weight}
\end{figure*}



\begin{table}[]
\begin{center}

\begin{tabular}{l|lll}
\hline
iMiGUE           & AMR 1 & AMR 2 & AMR 3 \\ \hline
Mean value of $W_{R}$ & 0.4666   & 0.4663   & 0.4665   \\
Mean value of $W_{s}$ & 0.4693   & 0.4939   & 0.4905   \\ \hline
NTURGB+D         & AMR 1 & AMR 2 & AMR 3 \\ \hline
Mean value of $W_{R}$ & 0.3067   & 0.3065   & 0.3067   \\
Mean value of $W_{s}$ & 0.3187   & 0.3275   & 0.3240   \\ \hline
\end{tabular}
\end{center}
\caption{The mean weights of different AMRs trained on the iMiGUE and NTURGB+D datasets.}
\label{tab:weight}
\end{table}

\begin{table}[]
\begin{center}
\resizebox{\columnwidth}{!}{    
\begin{tabular}{c|c|cccc|c}
\hline
Defence(NTU-RGB+D)     & \ding{51}RS             & \ding{55}R              & \ding{55}S              & \ding{55}RS             & Avg            & RI             \\ \hline
None                  & \textbf{88.09} & 0.98           & 9.00           & 0.00           & 3.33           & 0              \\
AT with \ding{55}RS(0 AMR) & 61.42          & 61.41          & 38.74          & 38.82          & 46.32          & 16.32          \\
1 AMR              & 76.44          & \textbf{76.44} & 37.89          & 37.83          & 50.72          & 37.74          \\
2 AMRs             & 81.61          & 67.93          & \textbf{80.53} & 66.00          & 71.49          & 61.68          \\
3 AMRs             & 81.65          & 68.62          & 80.35          & \textbf{66.61} & \textbf{71.86} & \textbf{62.09} \\ \hline
Defence(iMiGUE)       & \ding{51}RS             & \ding{55}R              & \ding{55}S              & \ding{55}RS             & Avg            & RI             \\ \hline
None                  & \textbf{62.37} & 0.13           & 6.67           & 0.02           & 2.27           & 0              \\
AT with \ding{55}RS(0 AMR) & 39.01          & 39.01          & 30.94          & 30.97          & 33.64          & 8.01           \\
1 AMR              & 43.89          & \textbf{43.89} & 32.46          & 32.50          & 36.28          & \textbf{15.53} \\
2 AMRs             & 42.93          & 42.93          & 31.27          & 31.16          & 35.12          & 13.41          \\
3 AMRs             & 42.38          & 42.38          & \textbf{33.33} & \textbf{33.42} & \textbf{36.38} & 14.12          \\ \hline
\end{tabular}}
\end{center}
\caption{The impact of the number of AMRs on accuracy against CW (perturbation budgets $\epsilon_{R},\epsilon_{s} = 8/255$, step size $\alpha = 2/255$, optimized by PGD30) attack on both datasets.}
\label{tab:5}
\end{table}
\begin{table}
\begin{center}
\resizebox{\columnwidth}{!}{   
\begin{tabular}{c|c|cccc|c}
\hline
$\lambda$ & \ding{51}RS             & \ding{55}R              & \ding{55}S              & \ding{55}RS             & Avg            & RI             \\ \hline
None  & \textbf{62.37} & 0.11           & 6.97           & 0.07           & 2.38           & 0              \\
0.1   & 60.25          & 36.23          & \textbf{59.15} & 32.02          & \textbf{42.47} & \textbf{37.97} \\
0.5   & 45.08          & \textbf{45.08} & 27.39          & 27.24          & 33.24          & 13.57          \\
1     & 43.89          & 43.89          & 32.85          & 32.94          & 36.56          & 15.70          \\
2     & 42.76          & 42.76          & 34.82          & 34.80          & 37.46          & 15.47          \\
5     & 41.64          & 41.64          & 37.50          & \textbf{37.56} & 38.9           & 15.79          \\ \hline
\end{tabular}}
\end{center}
\caption{Robust comparison of different $\lambda$ on iMiGUE with 1 AMR.}
\label{tab:lambda}
\end{table}
\subsection{Weight Distribution Visualization}
As mentioned in \cref{sec:3.3}, our AMR can assign appropriate weights to the features of two modalities. Specifically, more robust features will receive higher weights, while less robust ones will receive smaller weights. To verify if our results align with the proposed idea, we visualize the attention weight distribution of the AMR in the pretrained network, denoted as $W_{R}$ and $W_{s}$ in the \cref{eq:matric}, as shown in the \cref{fig:weight}. 
Further, we also calculate the overall mean of weight matrices within each AMR. The results in \cref{tab:weight} indicate that in each AMR, the average weight values corresponding to the skeletal modality are greater than the average weight values for the RGB modality.The experimental results align with our expectations. Observing \cref{fig:weight}, we notice a more substantial fluctuation in the weight distribution of the skeleton modality compared to the RGB modality. This phenomenon can be attributed to the greater impact of adversarial attacks on the RGB modality. Additionally, the skeleton encompasses a broader range of motion features, leading the model to incline towards adjusting the weights of the skeleton modality during the training process to enhance overall robustness.

\subsection{Ablation Study}

\subsubsection{The Impact of the number of AMR}

Our proposed AMR is an AT-based method. In order to evaluate the effectiveness of AMR and investigate the impact of the number of AMRs on overall performance, we conduct ablation experiments using None and AT with \ding{55}RS (which can be considered as having no AMRs, \emph{i.e.}, 0 AMR) as baselines. We start from the last FC layer of the branch network and insert 1 to 3 AMRs in a forward manner. The other parameters remain the same as described in \cref{sec:4.2}. The results in \cref{tab:5} demonstrate that, with an increasing number of AMRs, the RI continuously improves on the NTU-RGB+D dataset. On the iMiGUE dataset, the RI reaches its peak with one AMR, and further increasing the number of AMRs leads to a decrease in RI. This is because NTU-RGB+D is a large-scale dataset containing 57k samples, while iMiGUE has only 18k samples. We can conclude that for larger datasets, we require more AMRs to achieve better performance. However, regardless of the number of AMRs, the use of AMR consistently outperforms not using AMR by a significant margin.

\subsubsection{The Impact of the hyperparameter $\lambda$}
In this part, we evaluate the impact of $\lambda$ in \cref{eq:total}. $\lambda$ acts as a hyperparameter, enabling us to fine-tune the trade-off between the two loss component. We train our models with one AMR on the iMiGUE dataset with 5 different values $\lambda = [0.1,0.5,1,2,5]$, and evaluate their accuracy under PGD20 attacks. The natural accuracy decrease with the increase of $\lambda$. This is because the increase of $\lambda$ reduces the proportion of natural samples in the loss function. The highest RI is achieved when $\lambda = 0.1$.

\section{Conclusion}
In this paper, we investigate the differences in robustness among various data modalities and find that the skeleton modality is more robust than the RGB. Based on this observation, we creatively propose the Attention-based Modality Reweighter (AMR). It autonomously reweights different modalities, enhancing overall robustness. To accommodate AMR, we introduce a new loss function that encompasses multiple auxiliary classification results, serving as deep supervision while considering both clean accuracy and robust accuracy. We conduct extensive experiments on popular benchmark datasets and evaluate performance against SOTA attack methods. The results demonstrate that our approach significantly improves the robustness of multimodal action recognition models compared to other methods. Importantly, it balances the discrepancy in robustness between different modalities and achieves new SOAT performance without any additional data.

\section{Acknowledement}
This work was supported by National Natural Science Foundation of China under Grant 62171309 and 62306061, and Guangdong Basic and Applied Basic Research Foundation (Grant No. 2023A1515140037).



{\small
\bibliographystyle{ieee}
\bibliography{main}

\begin{thebibliography}{10}\itemsep=-1pt

\bibitem{Andriushchenko2019SquareAA}
M.~Andriushchenko, F.~Croce, N.~Flammarion, and M.~Hein.
\newblock Square attack: A query-efficient black-box adversarial attack via random search.
\newblock In {\em Computer Vision – ECCV 2020: 16th European Conference, Glasgow, UK, August 23–28, 2020, Proceedings, Part XXIII}, page 484–501, Berlin, Heidelberg, 2020. Springer-Verlag.

\bibitem{Buch2018ArtificialII}
V.~H. Buch, I.~Ahmed, and M.~Maruthappu.
\newblock Artificial intelligence in medicine: current trends and future possibilities.
\newblock {\em British Journal of General Practice}, 68(668):143--144, 2018.

\bibitem{2016Towards}
N.~Carlini and D.~Wagner.
\newblock Towards evaluating the robustness of neural networks.
\newblock In {\em 2017 IEEE Symposium on Security and Privacy (SP)}, pages 39--57, 2017.

\bibitem{2017Quo}
J.~Carreira and A.~Zisserman.
\newblock Quo vadis, action recognition? a new model and the kinetics dataset.
\newblock In {\em 2017 IEEE Conference on Computer Vision and Pattern Recognition (CVPR)}, pages 4724--4733, 2017.

\bibitem{Chen2019ModelfreeDR}
J.~Chen, B.~Yuan, and M.~Tomizuka.
\newblock Model-free deep reinforcement learning for urban autonomous driving.
\newblock In {\em 2019 IEEE Intelligent Transportation Systems Conference (ITSC)}, pages 2765--2771, 2019.

\bibitem{Croce2019MinimallyDA}
F.~Croce and M.~Hein.
\newblock Minimally distorted adversarial examples with a fast adaptive boundary attack.
\newblock In {\em Proceedings of the 37th International Conference on Machine Learning}, ICML'20. JMLR.org, 2020.

\bibitem{Croce2020ReliableEO}
F.~Croce and M.~Hein.
\newblock Reliable evaluation of adversarial robustness with an ensemble of diverse parameter-free attacks.
\newblock ICML'20, pages 2206--2216. JMLR.org, 2020.

\bibitem{Duan2021RevisitingSA}
H.~Duan, Y.~Zhao, K.~Chen, D.~Lin, and B.~Dai.
\newblock Revisiting skeleton-based action recognition.
\newblock In {\em 2022 IEEE/CVF Conference on Computer Vision and Pattern Recognition (CVPR)}, pages 2959--2968, 2022.

\bibitem{duan2022novel}
M.~Duan, K.~Li, J.~Deng, B.~Xiao, and Q.~Tian.
\newblock A novel multi-sample generation method for adversarial attacks.
\newblock {\em ACM Trans. Multimedia Comput. Commun. Appl.}, 18(4), mar 2022.

\bibitem{Eykholt2018RobustPA}
K.~Eykholt, I.~Evtimov, E.~Fernandes, B.~Li, A.~Rahmati, C.~Xiao, A.~Prakash, T.~Kohno, and D.~Song.
\newblock Robust physical-world attacks on deep learning visual classification.
\newblock In {\em Proceedings of the IEEE conference on computer vision and pattern recognition}, pages 1625--1634, 2018.

\bibitem{Fu2023LearningSR}
Z.~Fu, Z.~Mao, Y.~Song, and Y.~Zhang.
\newblock Learning semantic relationship among instances for image-text matching.
\newblock In {\em 2023 IEEE/CVF Conference on Computer Vision and Pattern Recognition (CVPR)}, pages 15159--15168, 2023.

\bibitem{2009Super}
D.~Glasner, S.~Bagon, and M.~Irani.
\newblock Super-resolution from a single image.
\newblock In {\em 2009 IEEE 12th international conference on computer vision}, pages 349--356. IEEE, 2009.

\bibitem{Goodfellow2014ExplainingAH}
I.~Goodfellow, J.~Shlens, and C.~Szegedy.
\newblock Explaining and harnessing adversarial examples.
\newblock {\em CoRR}, abs/1412.6572, 2014.

\bibitem{He2015DeepRL}
K.~He, X.~Zhang, S.~Ren, and J.~Sun.
\newblock Deep residual learning for image recognition.
\newblock In {\em 2016 IEEE Conference on Computer Vision and Pattern Recognition (CVPR)}, pages 770--778, 2016.

\bibitem{Kannan2018AdversarialLP}
H.~Kannan, A.~Kurakin, and I.~Goodfellow.
\newblock Adversarial logit pairing.
\newblock {\em ArXiv}, abs/1803.06373, 2018.

\bibitem{Khader2022MedicalDW}
F.~Khader, G.~Mueller-Franzes, T.~Wang, T.~Han, S.~T. Arasteh, C.~Haarburger, J.~Stegmaier, K.~Bressem, C.~Kuhl, S.~Nebelung, J.~N. Kather, and D.~Truhn.
\newblock Medical diagnosis with large scale multimodal transformers: Leveraging diverse data for more accurate diagnosis.
\newblock {\em ArXiv}, abs/2212.09162, 2022.

\bibitem{Kinfu2022AnalysisAE}
K.~A. Kinfu and R.~Vidal.
\newblock Analysis and extensions of adversarial training for video classification.
\newblock {\em 2022 IEEE/CVF Conference on Computer Vision and Pattern Recognition Workshops (CVPRW)}, pages 3415--3424, 2022.

\bibitem{Kong2017CancerMD}
B.~Kong, X.~Wang, Z.~Li, Q.~Song, and S.~Zhang.
\newblock Cancer metastasis detection via spatially structured deep network.
\newblock In M.~Niethammer, M.~Styner, S.~Aylward, H.~Zhu, I.~Oguz, P.-T. Yap, and D.~Shen, editors, {\em Information Processing in Medical Imaging}, pages 236--248, Cham, 2017. Springer International Publishing.

\bibitem{Kullback1951OnIA}
S.~Kullback and R.~A. Leibler.
\newblock On information and sufficiency.
\newblock {\em Annals of Mathematical Statistics}, 22:79--86, 1951.

\bibitem{2018Co}
C.~Li, Q.~Zhong, D.~Xie, and S.~Pu.
\newblock Co-occurrence feature learning from skeleton data for action recognition and detection with hierarchical aggregation.
\newblock In {\em Proceedings of the 27th International Joint Conference on Artificial Intelligence}, IJCAI'18, page 786–792. AAAI Press, 2018.

\bibitem{2020BERT}
L.~Li, R.~Ma, Q.~Guo, X.~Xue, and X.~Qiu.
\newblock Bert-attack: Adversarial attack against bert using bert.
\newblock In {\em Proceedings of the 2020 Conference on Empirical Methods in Natural Language Processing (EMNLP)}, pages 6193--6202. Association for Computational Linguistics, 2020.

\bibitem{Liu2021iMiGUEAI}
X.~Liu, H.~Shi, H.~Chen, Z.~Yu, X.~Li, and G.~Zhao.
\newblock imigue: An identity-free video dataset for micro-gesture understanding and emotion analysis.
\newblock In {\em 2021 IEEE/CVF Conference on Computer Vision and Pattern Recognition (CVPR)}, pages 10626--10637, 2021.

\bibitem{liu2023multi}
X.~Liu, K.~Yuan, X.~Niu, J.~Shi, Z.~Yu, H.~Yue, and J.~Yang.
\newblock Multi-scale promoted self-adjusting correlation learning for facial action unit detection.
\newblock {\em arXiv preprint arXiv:2308.07770}, 2023.

\bibitem{liu2024rppg}
X.~Liu, Y.~Zhang, Z.~Yu, H.~Lu, H.~Yue, and J.~Yang.
\newblock rppg-mae: Self-supervised pretraining with masked autoencoders for remote physiological measurements.
\newblock {\em IEEE Transactions on Multimedia}, 2024.

\bibitem{lu2024gpt}
H.~Lu, X.~Niu, J.~Wang, Y.~Wang, Q.~Hu, J.~Tang, Y.~Zhang, K.~Yuan, B.~Huang, Z.~Yu, et~al.
\newblock Gpt as psychologist? preliminary evaluations for gpt-4v on visual affective computing.
\newblock In {\em Proceedings of the IEEE/CVF Conference on Computer Vision and Pattern Recognition}, pages 322--331, 2024.

\bibitem{Ma2019UnderstandingAA}
X.~Ma, Y.~Niu, L.~Gu, Y.~Wang, Y.~Zhao, J.~Bailey, and F.~Lu.
\newblock Understanding adversarial attacks on deep learning based medical image analysis systems.
\newblock {\em Pattern Recognition}, 110:107332, 2021.

\bibitem{2019Adversarial}
D.~Madaan, J.~Shin, and S.~J. Hwang.
\newblock Adversarial neural pruning with latent vulnerability suppression.
\newblock In {\em Proceedings of the 37th International Conference on Machine Learning}, ICML'20, pages 6575--6585. JMLR.org, 2020.

\bibitem{2017Towards}
A.~Madry, A.~Makelov, L.~Schmidt, D.~Tsipras, and A.~Vladu.
\newblock Towards deep learning models resistant to adversarial attacks.
\newblock In {\em 6th International Conference on Learning Representations, {ICLR} 2018, Vancouver, BC, Canada, April 30 - May 3, 2018, Conference Track Proceedings}. OpenReview.net, 2018.

\bibitem{Moon2021MultiModalUA}
J.~H. Moon, H.~Lee, W.~Shin, Y.-H. Kim, and E.~Choi.
\newblock Multi-modal understanding and generation for medical images and text via vision-language pre-training.
\newblock {\em IEEE Journal of Biomedical and Health Informatics}, 26(12):6070--6080, 2022.

\bibitem{2015DeepFool}
S.-M. Moosavi-Dezfooli, A.~Fawzi, and P.~Frossard.
\newblock Deepfool: A simple and accurate method to fool deep neural networks.
\newblock In {\em 2016 IEEE Conference on Computer Vision and Pattern Recognition (CVPR)}, pages 2574--2582, 2016.

\bibitem{1993A}
N.~R. Pal and S.~K. Pal.
\newblock A review on image segmentation techniques.
\newblock {\em Pattern recognition}, 26(9):1277--1294, 1993.

\bibitem{2016You}
J.~Redmon, S.~Divvala, R.~Girshick, and A.~Farhadi.
\newblock You only look once: Unified, real-time object detection.
\newblock In {\em Proceedings of the IEEE conference on computer vision and pattern recognition}, pages 779--788, 2016.

\bibitem{Shahroudy2016NTURA}
A.~Shahroudy, J.~Liu, T.-T. Ng, and G.~Wang.
\newblock Ntu rgb+d: A large scale dataset for 3d human activity analysis.
\newblock In {\em 2016 IEEE Conference on Computer Vision and Pattern Recognition (CVPR)}, pages 1010--1019, 2016.

\bibitem{Su2017OnePA}
J.~Su, D.~V. Vargas, and K.~Sakurai.
\newblock One pixel attack for fooling deep neural networks.
\newblock {\em IEEE Transactions on Evolutionary Computation}, 23(5):828--841, 2019.

\bibitem{2013Intriguing}
C.~Szegedy, W.~Zaremba, I.~Sutskever, J.~Bruna, D.~Erhan, I.~Goodfellow, and R.~Fergus.
\newblock Intriguing properties of neural networks.
\newblock {\em arXiv preprint arXiv:1312.6199}, 2013.

\bibitem{Tian2021CanAI}
Y.~Tian and C.~Xu.
\newblock Can audio-visual integration strengthen robustness under multimodal attacks?
\newblock In {\em 2021 IEEE/CVF Conference on Computer Vision and Pattern Recognition (CVPR)}, pages 5597--5607, 2021.

\bibitem{Joze2019MMTMMT}
H.~R. Vaezi~Joze, A.~Shaban, M.~L. Iuzzolino, and K.~Koishida.
\newblock Mmtm: Multimodal transfer module for cnn fusion.
\newblock In {\em 2020 IEEE/CVF Conference on Computer Vision and Pattern Recognition (CVPR)}, pages 13286--13296, 2020.

\bibitem{2020Transferable}
H.~Wang, G.~Wang, Y.~Li, D.~Zhang, and L.~Lin.
\newblock Transferable, controllable, and inconspicuous adversarial attacks on person re-identification with deep mis-ranking.
\newblock In {\em Proceedings of the IEEE/CVF conference on computer vision and pattern recognition}, pages 342--351, 2020.

\bibitem{Wang2023ADS}
L.~Wang, Z.~He, J.~Tang, R.~Dang, N.~Wang, C.~Liu, and Q.~Chen.
\newblock A dual semantic-aware recurrent global-adaptive network for vision-and-language navigation.
\newblock In E.~Elkind, editor, {\em Proceedings of the Thirty-Second International Joint Conference on Artificial Intelligence, {IJCAI-23}}, pages 1479--1487. International Joint Conferences on Artificial Intelligence Organization, 8 2023.
\newblock Main Track.

\bibitem{Wang2020ImprovingAR}
Y.~Wang, D.~Zou, J.~Yi, J.~Bailey, X.~Ma, and Q.~Gu.
\newblock Improving adversarial robustness requires revisiting misclassified examples.
\newblock In {\em 8th International Conference on Learning Representations, {ICLR} 2020, Addis Ababa, Ethiopia, April 26-30, 2020}. OpenReview.net, 2020.

\bibitem{2018Sparse}
X.~Wei, J.~Zhu, S.~Yuan, and H.~Su.
\newblock Sparse adversarial perturbations for videos.
\newblock In {\em Proceedings of the AAAI Conference on Artificial Intelligence}, volume~33, pages 8973--8980, 2019.

\bibitem{wu2023adaptive}
S.~Wu, J.~Sang, K.~Xu, G.~Zheng, and C.~Xu.
\newblock Adaptive adversarial logits pairing.
\newblock {\em ACM Transactions on Multimedia Computing, Communications and Applications}, 20(2), oct 2023.

\bibitem{xu2023a2sc}
Y.~Xu, X.~Wei, P.~Dai, and X.~Cao.
\newblock A2sc: Adversarial attacks on subspace clustering.
\newblock {\em ACM Trans. Multimedia Comput. Commun. Appl.}, 19(6), jul 2023.

\bibitem{2018Spatial}
S.~Yan, Y.~Xiong, and D.~Lin.
\newblock Spatial temporal graph convolutional networks for skeleton-based action recognition.
\newblock In {\em Proceedings of the Thirty-Second AAAI Conference on Artificial Intelligence and Thirtieth Innovative Applications of Artificial Intelligence Conference and Eighth AAAI Symposium on Educational Advances in Artificial Intelligence}, AAAI'18/IAAI'18/EAAI'18. AAAI Press, 2018.

\bibitem{Yu2022MMNetAM}
B.~X. Yu, Y.~Liu, X.~Zhang, S.-h. Zhong, and K.~C. Chan.
\newblock Mmnet: A model-based multimodal network for human action recognition in rgb-d videos.
\newblock {\em IEEE Transactions on Pattern Analysis and Machine Intelligence}, 45(3):3522--3538, 2023.

\bibitem{yuan2024auformer}
K.~Yuan, Z.~Yu, X.~Liu, W.~Xie, H.~Yue, and J.~Yang.
\newblock Auformer: Vision transformers are parameter-efficient facial action unit detectors.
\newblock {\em arXiv preprint arXiv:2403.04697}, 2024.

\bibitem{Zhang2019TheoreticallyPT}
H.~Zhang, Y.~Yu, J.~Jiao, E.~Xing, L.~E. Ghaoui, and M.~Jordan.
\newblock Theoretically principled trade-off between robustness and accuracy.
\newblock In K.~Chaudhuri and R.~Salakhutdinov, editors, {\em Proceedings of the 36th International Conference on Machine Learning}, volume~97 of {\em Proceedings of Machine Learning Research}, pages 7472--7482. PMLR, 09--15 Jun 2019.

\bibitem{2020Geometry}
J.~Zhang, J.~Zhu, G.~Niu, B.~Han, M.~Sugiyama, and M.~S. Kankanhalli.
\newblock Geometry-aware instance-reweighted adversarial training.
\newblock In {\em 9th International Conference on Learning Representations, {ICLR} 2021, Virtual Event, Austria, May 3-7, 2021}. OpenReview.net, 2021.

\bibitem{zhang2024advancing}
Y.~Zhang, H.~Lu, X.~Liu, Y.~Chen, and K.~Wu.
\newblock Advancing generalizable remote physiological measurement through the integration of explicit and implicit prior knowledge.
\newblock {\em arXiv preprint arXiv:2403.06947}, 2024.

\end{thebibliography}
}

\clearpage
\setcounter{page}{1}
\maketitlesupplementary

\begin{table}[]
\begin{center}
\resizebox{\columnwidth}{!}{
\begin{tabular}{c|c|cccc|c}
\hline
Defence (NTURGB+D)   & \ding{51}RS             & \ding{55}R              & \ding{55}S              & \ding{55}RS             & Avg            & RI             \\ \hline
None                & \textbf{88.09} & 49.08          & 64.05          & 17.36          & 43.50          & 0              \\
AT\cite{2017Towards} with \ding{55}R    & 75.59          & \textbf{75.59} & 22.18          & 22.20          & 39.99          & -16.01         \\
AT\cite{2017Towards} with \ding{55}S    & 78.44          & 8.36           & 78.35          & 8.33           & 31.68          & -21.47         \\
AT\cite{2017Towards} with \ding{55}RS & 61.42          & 61.41          & 47.04          & 47.00          & 51.82          & -18.35         \\
MinSim\cite{Tian2021CanAI}              & 80.85          & 13.80          & 80.85          & 13.80          & 36.15          & -14.59         \\
ExFMem\cite{Tian2021CanAI}              & 84.19          & 20.96          & 75.27          & 10.59          & 35.61          & -11.79         \\
MinSim+ExFMem\cite{Tian2021CanAI}       & 82.81          & 17.39          & 80.27          & 16.25          & 37.97          & -10.81         \\
AMR(Ours)           & 81.65          & 71.99          & \textbf{80.77} & \textbf{70.63} & \textbf{74.46} & \textbf{24.97} \\ \hline
Defence (iMiGUE)     & \ding{51}RS             & \ding{55}R              & \ding{55}S              & \ding{55}RS             & Avg            & RI             \\ \hline
None                & \textbf{62.37} & 5.98           & 48.85          & 2.26           & 19.03          & 0              \\
AT\cite{2017Towards} with \ding{55}R    & 47.32          & \textbf{47.32} & 10.54          & 9.77           & 22.54          & -11.54         \\
AT\cite{2017Towards} with \ding{55}S    & 60.20          & 2.78           & \textbf{60.11} & 3.11           & 22.0           & 0.8            \\
AT\cite{2017Towards} with \ding{55}RS & 39.01          & 38.98          & 34.43          & 34.36          & 35.92          & -6.47          \\
MinSim\cite{Tian2021CanAI}              & 60.60          & 5.19           & 57.59          & 3.62           & 22.13          & \textbf{1.33}  \\
ExFMem\cite{Tian2021CanAI}              & 60.53          & 6.14           & 55.64          & 2.98           & 21.59          & 0.72           \\
MinSim+ExFMem\cite{Tian2021CanAI}       & 60.62          & 4.30           & 58.36          & 3.29           & 21.98          & 1.2            \\
AMR(Ours)           & 43.89          & 43.89          & 36.18          & \textbf{36.45} & \textbf{38.84} & 1.17           \\ \hline
\end{tabular}}
\end{center}
\caption{The various accuracy(\%) metrics on the NTU-RGB+D and iMiGUE datasets are presented with different defense methods. Here, we utilize FGSM (perturbation budgets $\epsilon_{R},\epsilon_{s} = 8/255$,step size $\alpha = 2/255$) to generate adversarial samples for assessing robustness accuracy. The best results among the metrics are denoted in bold.}
\label{tab:FGSM attack}
\end{table}

\begin{table}[]
\begin{center}
\resizebox{\columnwidth}{!}{
\begin{tabular}{c|c|cccc|c}
\hline
Defence(NTURGB+D) & \ding{51}RS             & \ding{55}R              & \ding{55}S              & \ding{55}RS             & Avg            & RI             \\ \hline
None              & \textbf{88.09} & 0.98           & 9.00           & 0.00           & 3.33           & 0              \\
AT\cite{2017Towards} with \ding{55}R        & 75.59          & \textbf{75.59} & 2.80           & 2.89           & 27.09          & 11.26          \\
AT\cite{2017Towards} with \ding{55}S        & 78.44          & 1.41           & 78.31          & 1.41           & 27.04          & 14.06          \\
AT\cite{2017Towards} with \ding{55}RS       & 61.42          & 61.41          & 38.74          & 38.82          & 46.32          & 16.32          \\
MinSim\cite{Tian2021CanAI}            & 80.85          & 0.50           & 80.84          & 0.48           & 27.27          & 16.70          \\
ExFMem\cite{Tian2021CanAI}            & 84.19          & 0.00           & 60.98          & 0.00           & 20.33          & 13.10          \\
MinSim+ExFMem\cite{Tian2021CanAI}     & 82.81          & 0.00           & 75.60          & 0.00           & 25.20          & 16.59          \\
AMR(Ours)      & 81.65          & 68.62          & \textbf{80.35} & \textbf{66.61} & \textbf{71.86} & \textbf{62.09} \\ \hline
Defence(iMiGUE)   & \ding{51}RS             & \ding{55}R              & \ding{55}S              & \ding{55}RS             & Avg            & RI             \\ \hline
None              & \textbf{62.37} & 0.13           & 6.67           & 0.02           & 2.27           & 0              \\
AT\cite{2017Towards} with \ding{55}R        & 47.32          & \textbf{47.32} & 0.00           & 0.00           & 15.77          & -1.55          \\
AT\cite{2017Towards} with \ding{55}S        & 60.20          & 0.00           & \textbf{59.79} & 0.00           & 19.93          & 15.49 \\
AT\cite{2017Towards} with \ding{55}RS       & 39.01          & 39.01          & 30.94          & 30.97          & 33.64          & 8.01           \\
MinSim\cite{Tian2021CanAI}            & 60.60          & 0.00           & 53.74          & 0.02           & 17.92          & 13.88          \\
ExFMem\cite{Tian2021CanAI}            & 60.53          & 0.00           & 45.19          & 0.00           & 15.06          & 10.95          \\
MinSim+ExFMem\cite{Tian2021CanAI}     & 60.62          & 0.00           & 54.72          & 0.00           & 18.24          & 14.20          \\
AMR(Ours)      & 43.89          & 43.89          & 32.46          & \textbf{32.50} & \textbf{36.28} & \textbf{15.53}          \\ \hline
\end{tabular}}
\end{center}
\caption{The various accuracy(\%) metrics on the NTU-RGB+D and iMiGUE datasets are presented with different defense methods. Here, we utilize CW attack (perturbation budgets $\epsilon_{R},\epsilon_{s} = 8/255$, step size $\alpha = 2/255$, optimized by PGD30) to generate adversarial samples for assessing robustness accuracy. The best results among the metrics are denoted in bold.}
\label{tab:3}
\end{table}

\begin{table}[]
\begin{center}
\resizebox{\columnwidth}{!}{
\begin{tabular}{c|c|cccc|c}
\hline
Defence(NTURGB+D)     & \ding{51}RS             & \ding{55}R              & \ding{55}S              & \ding{55}RS             & Avg            & RI             \\ \hline
None                  & \textbf{88.09} & 2.09           & 11.84          & 0.15           & 4.69           & 0              \\
AT with \ding{55}RS(0 AMR) & 61.42          & 61.41          & 38.74          & 38.82          & 47.38          & 16.02          \\
1 AMR              & 76.44          & \textbf{76.44} & 38.29          & 38.32          & 51.02          & 34.68          \\
2 AMRs             & 81.61          & 67.80          & \textbf{80.58} & 65.88          & 71.42          & 60.25          \\
3 AMRs             & 81.65          & 68.54          & 80.40          & \textbf{66.52} & \textbf{71.82} & \textbf{60.69} \\ \hline
Defence(iMiGUE)       & \ding{51}RS             & \ding{55}R              & \ding{55}S              & \ding{55}RS             & Avg            & RI             \\ \hline
None                  & \textbf{62.37} & 0.13           & 6.67           & 0.02           & 2.27           & 0              \\
AT with \ding{55}RS(0 AMR) & 39.01          & 39.01          & 31.89          & 31.95          & 34.28          & 8.54           \\
1 AMR              & 43.89          & \textbf{43.89} & 32.85          & 32.94          & 36.56          & \textbf{15.70} \\
2 AMRs             & 42.93          & 42.93          & 32.35          & 32.43          & 35.92          & 14.08          \\
3 AMRs             & 42.38          & 42.38          & \textbf{34.34} & \textbf{34.32} & \textbf{37.01} & 14.64          \\ \hline
\end{tabular}}
\end{center}
\caption{The impact of the number of AMRs on accuracy against PGD20 (perturbation budgets $\epsilon_{R},\epsilon_{s} = 8/255$, step size $\alpha = 2/255$) attack on both datasets.}
\label{tab:4}
\end{table}

\end{document}